\UseRawInputEncoding
\documentclass[preprint,12pt]{elsarticle}
\usepackage{amssymb}
\usepackage{placeins}
\usepackage{colortbl}
\usepackage{amsmath}
\usepackage{multirow}
\usepackage{pdfpages}
\usepackage{algorithm}
\usepackage{algorithmic}

\biboptions{sort&compress}

\journal{Applied Soft Computing}

\begin{document}

\begin{frontmatter}

\title{Synaptic Pruning: A Biological Inspiration for Deep Learning Regularization}
\author[inst1]{Gideon Vos}

\affiliation[inst1]{organization={College of Science and Engineering, James Cook University},
            addressline={James Cook Dr}, 
            city={Townsville},
            postcode={4811}, 
            state={QLD},
            country={Australia}}

\author[inst2]{Liza van Eijk}
\author[inst3]{Zoltan Sarnyai}
\author[inst1]{Mostafa Rahimi Azghadi}

\affiliation[inst2]{organization={College of Health Care Sciences, James Cook University},
            addressline={James Cook Dr}, 
            city={Townsville},
            postcode={4811}, 
            state={QLD},
            country={Australia}}

\affiliation[inst3]{organization={College of Public Health, Medical, and Vet Sciences, James Cook University},
            addressline={James Cook Dr}, 
            city={Townsville},
            postcode={4811}, 
            state={QLD},
            country={Australia}}
\begin{abstract}
Biological synaptic pruning removes weak neural connections to improve efficiency, while standard dropout in artificial networks randomly deactivates neurons without considering connection importance. We propose a magnitude-based synaptic pruning method that better emulates biological processes by gradually removing connections according to their contribution to model performance. Integrated directly into the training loop as a dropout replacement, our method computes weight importance from absolute magnitudes across layers and applies a cubic schedule to progressively increase global sparsity. At regular intervals, pruning masks are updated by thresholding weights, permanently removing low-importance connections while preserving gradient flow for active ones. This continuous, data-driven pruning removes the need for separate pruning and fine-tuning phases. We evaluated the method across multiple time series forecasting architectures, including Recurrent Neural Networks, Long Short-Term Memory, and Patch Time Series Transformer models, using four datasets. Our synaptic pruning approach achieved the best overall performance ranking across all architectures, with statistically significant improvements confirmed by Friedman tests ($p < 0.01$). In financial forecasting tasks, it reduced Mean Absolute Error by up to 20\% compared to models using no dropout or standard dropout, with reductions reaching 52\% in select transformer models. The proposed mechanism advances regularization by coupling dynamic weight elimination with progressive sparsification during training. Its ease of integration and consistent accuracy gains make it a practical alternative to dropout, particularly for high-precision domains such as financial time series forecasting. Source code is available at https://github.com/xalentis/SynapticPruning

\end{abstract}

\begin{graphicalabstract}
\begin{center}
  \makebox[\textwidth]{\includegraphics[width=\paperwidth]{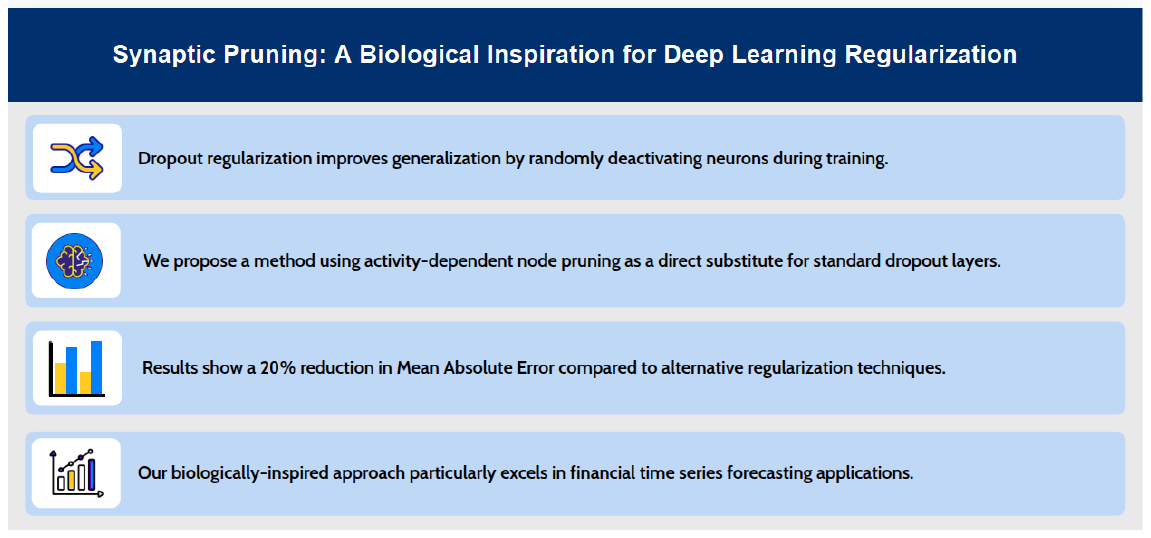}}
\end{center}
\end{graphicalabstract}

\begin{highlights}
\item We propose a magnitude-based synaptic pruning method as a regularization approach
\item The method emulates biological synaptic pruning by dynamically and progressively eliminating low-importance neural connections
\item Experiments across RNN, LSTM, and PatchTST models show statistically significant error reductions
\item Our approach reduces predictive error rates by up to 52\% in times series forecasting tasks
\end{highlights}


\end{frontmatter}



\section{Introduction}
\noindent Synaptic pruning is a fundamental neuro-developmental process whereby weak or redundant synaptic connections are selectively eliminated to enhance the efficiency of neuronal networks \cite{rakic1986}. This process is highly activity-dependent, meaning that synapses with lower activity or utility are more likely to be pruned, resulting in improved energy efficiency and functional specialization \cite{Huttenlocher1997}. During critical periods of development, the human brain initially overproduces synapses, creating a dense network of connections that undergoes substantial refinement through adolescence and early adulthood \cite{Chechik1998, Selemon1999}. This refinement is essential for optimal cognitive function and normal brain development, with disruptions in pruning mechanisms associated with various neuro-developmental disorders \cite{Zhan2014, Tang2014}. \\

\noindent In the field of artificial neural networks, researchers have drawn inspiration from biological synaptic pruning to develop methods for improving network efficiency and generalization capacity \cite{LeCun1990, Hassabis2017, Han2015}. Dropout regularization \cite{Srivastava2014} functions by randomly deactivating, or "dropping out," a subset of neurons during each training iteration. This process forces the network to avoid over-reliance on any particular neuron, promoting redundancy and robustness in the learned representations. By preventing neurons from co-adapting too strongly to specific patterns in the training data, dropout reduces the risk of overfitting. As a result, the model is encouraged to generalize better to unseen data, leading to improved performance during inference. More recent approaches have extended beyond dropout to implement more biologically-inspired pruning methods that permanently remove weights or connections based on importance criteria \cite{Han2015, Molchanov2017}, potentially offering advantages in both computational efficiency and performance. \\

\noindent In this work, we introduce a novel regularization method that more closely emulates biological synaptic pruning within deep neural networks, with a particular focus on enhancing time series forecasting performance. Unlike standard dropout, which randomly and temporarily deactivates neurons during training, or pruning techniques aimed solely at post-training model compression, our method applies an activity-dependent, gradual elimination of connections based on their contribution to model performance \cite{Frankle2019}. By combining temporal scheduling with context-sensitive pruning, our approach progressively weakens and permanently removes low-importance connections during training, enabling the network to adapt dynamically. By integrating directly into the training process, our method requires no additional pruning or fine-tuning stages and is especially effective in recurrent and temporal models commonly used for time series forecasting.

\section{Related Works}
\noindent In computational neuroscience and deep learning, several methods have been proposed to emulate biological synaptic pruning. LeCun \emph{et al.} proposed Optimal Brain Damage \cite{LeCun1990}, one of the earliest approaches to biologically inspired pruning in artificial neural networks. This method uses second-order derivatives of the loss function to estimate the saliency of each weight, pruning those with minimal impact on performance. However, this approach requires complete model training before pruning begins, making it a post-training optimization rather than an integrated learning process that could adapt to evolving temporal patterns in time series data.\\

\noindent Han et al. \cite{Han2015} introduced magnitude-based pruning, where neural network weights with magnitudes below a certain threshold are eliminated, under the assumption that they contribute less to the final output. While effective in reducing network complexity, this approach primarily focuses on static weight values and does not consider the dynamic importance of individual nodes or features over time, which is particularly limiting for time series forecasting where temporal dependencies change throughout the learning process. This approach further requires three independent steps consisting of initial model training, pruning, and finally model fine-tuning, making it less practical for applications where continuous adaptation during training is valuable. \\

\noindent Li et al. \cite{Li2016} proposed sensitivity-based pruning for Convolutional Networks, also known as contribution-aware pruning, which evaluates the increase in loss when individual neurons are ablated, thereby preserving units that significantly impact performance. This method aligns with the biological concept of activity-dependent synaptic retention, where synapses that contribute more to network function are retained, while those with minimal impact are pruned. However, this approach still operates through discrete evaluation phases rather than continuous monitoring, which may miss the evolving importance of features during model training and inference. \\

\noindent Gradient-based pruning was proposed by Molchanov et al. \cite{Molchanov2017}, which analyzes the gradient flow through each neuron or connection and prunes those with consistently low gradient magnitudes, indicating minimal contribution to loss reduction. While this approach captures some dynamic behavior of the network, it often incurs high computational overhead due to the need to calculate gradients for every neuron or connection during training, and typically requires separate pruning phases distinct from regular training. \\

\noindent Hassabis et al. \cite{Hassabis2017} presented a broader framework linking neuroscience and artificial intelligence, emphasizing how mechanisms such as synaptic pruning and activity-dependent plasticity inform efficient computation in the brain. They argued that integrating biological principles into artificial systems, such as pruning underused or redundant connections, can lead to more robust, adaptable, and generalizable models. This perspective supports pruning not merely as a compression technique, but as a fundamental principle for learning and generalization, grounded in neuro-biological theory, which is particularly relevant for time series forecasting where adaptability to changing patterns is crucial. \\

\noindent Frankle et al. \cite{Frankle2019} introduced the Lottery Ticket Hypothesis (LTH), which suggests that within large networks exist sparse sub-networks, referred to as "winning tickets," that can achieve comparable performance if identified early and pruned effectively. LTH finds inherently trainable sparse architectures that require retraining from specific initial conditions. This approach mimics developmental pruning by identifying and retaining strong sub-networks early in training, much like how the brain strengthens key neural connections during development while pruning others. However, the iterative process of training, pruning, and retraining may be less practical for applications where continuous adaptation during a single training cycle is preferred. \\

\noindent Gradient-based Single-shot Pruning (SNIP) was introduced by Lee \emph{et al.} \cite{lee2019}, which estimates the importance of each connection at initialization by measuring the sensitivity of the loss to small perturbations in the weights. Connections with low gradient-based saliency scores are pruned before training begins, under the assumption that their contribution to learning will remain negligible. While computationally efficient, this approach makes one-time pruning decisions based on the network's initial state, assuming that early indicators reliably predict long-term relevance, which may not hold for cases where feature importance can evolve as the model learns different temporal patterns. \\

\noindent Wang \emph{et al.} \cite{wang2020} proposed Gradient Signal Preservation (GraSP), aiming to maintain train-ability in pruned networks by preserving the gradient flow at initialization. GraSP evaluates the influence of each weight on the network's Hessian-gradient product and prunes connections that minimally affect gradient propagation. This method improves over naive pruning by considering second-order interactions. However, like SNIP, it makes static pruning decisions at initialization rather than adapting to changing feature importance throughout training, which limits its effectiveness when temporal relationships may shift during the learning process. \\

\noindent In contrast to these existing approaches, our proposed method operates as a potential replacement or augmentation for standard dropout during training, continuously monitoring and updating connection importance based on their contribution to reducing prediction error. While preliminary experiments with computer vision architectures, including EfficientNet, DenseNet, and VGG, showed minimal efficacy, this dynamic, activity-dependent approach proved particularly well-suited for time series forecasting applications where temporal patterns and feature relevance can evolve as the model learns, and where the ability to identify and preserve the most predictive temporal relationships throughout training is crucial for achieving optimal performance.

\section{Methods}

\subsection{Proposed Pruning Method}

\noindent The method proposed in this study implements a global magnitude-based weight pruning strategy that progressively removes connections based on their importance to the network's predictive performance. Our synaptic pruning approach differs fundamentally from standard dropout \cite{Srivastava2014} and other regularization techniques in several key aspects. Specifically, our method:

\begin{itemize}
\item Employs global magnitude-based pruning across all network layers simultaneously
\item Implements a cubic scheduling function for gradual sparsity increase over training epochs
\item Maintains binary masks that permanently remove connections rather than stochastic deactivation
\item Uses a warmup period to allow initial weight development before pruning begins
\item Results in a structurally sparse network with reduced computational requirements
\end{itemize}

\noindent Our proposed method operates at the individual weight level using global magnitude comparison, ensuring that only the least important connections across the entire network are removed. This approach aligns well with biological synaptic pruning principles where weaker synaptic connections are eliminated to optimize neural efficiency \cite{chechik1999, lecun2021, zhu2017}.

\subsection{Pruning Components}

\noindent The synaptic pruning algorithm consists of three main components: i) mask initialization, ii) cubic sparsity scheduling, and iii) global magnitude-based selection.

\subsubsection{Mask Initialization}

\noindent Upon initialization, binary masks are created for all weight parameters in the fully connected layers. These masks are initialized as binary tensors.

\subsubsection{Cubic Sparsity Scheduling}

\noindent The target sparsity follows a cubic scheduling function that enables gradual pruning progression. After a warmup period of 2 epochs (adjustable as a parameter), the sparsity increases from a minimum rate of 30\% to a maximum of 70\% according to:

\[
s(t) = s_{\text{min}} + (s_{\text{max}} - s_{\text{min}}) \cdot \left( \frac{t_{\text{total}} - t_{\text{warmup}}}{t - t_{\text{warmup}}} \right)^3
\]

\noindent where $s(t)$ is the target sparsity at epoch $t$, $s_{\min} = 0.3$, $s_{\max} = 0.7$, $t_{\text{warmup}} = 2$, and $t_{\text{total}} = 20$.

\subsubsection{Global Magnitude-based Selection}

\noindent The core pruning mechanism evaluates all active (non-pruned) weights across the network simultaneously. At each pruning step, our method:

\begin{itemize}
    \item Collects all currently active weights from all modules
    \item Calculates the absolute magnitude of each weight
    \item Determines a global threshold based on the target number of weights to prune
    \item Updates masks to zero out weights below the threshold
    \item Applies the updated masks to preserve the pruned structure
\end{itemize}

\noindent This global selection ensures that pruning decisions consider the relative importance of weights across different layers, preventing any single layer from being over-pruned while others remain dense.

\subsection{Training Integration}

\noindent The proposed pruning method is integrated into the usual training loop of deep neural networks, with pruning applied every 5 batches (after the initial warmup period), allowing for frequent adaptation to changing weight importance. The pruning masks are applied both immediately after pruning updates and after each optimizer step to ensure pruned weights remain at zero. The six key algorithms utilized for synaptic pruning are shown as pseudo-code below, with the full implementation provided as Python source code available at https://github.com/xalentis/SynapticPruning.

\begin{algorithm}[H]
\caption{Synaptic Pruning Initialization}
\label{alg:synaptic_init}
\begin{algorithmic}[1]
\REQUIRE Neural network modules $\mathcal{M} = \{M_1, M_2, ..., M_k\}$
\REQUIRE Minimum sparsity rate $s_{min} = 0.3$
\REQUIRE Maximum sparsity rate $s_{max} = 0.7$
\REQUIRE Warmup epochs $t_{warmup} = 2$
\REQUIRE Total training epochs $t_{total} = 20$
\ENSURE Initialized binary masks $\mathcal{B}$

\STATE Initialize empty mask dictionary $\mathcal{B} \leftarrow \{\}$
\FOR{each module $M_i \in \mathcal{M}$}
    \FOR{each parameter $\theta \in M_i$ where $\theta$ is a weight tensor}
        \STATE Create binary mask $B_\theta \leftarrow \mathbf{1}_{|\theta|}$ \COMMENT{Initialize all weights as active}
        \STATE $\mathcal{B}[\theta] \leftarrow B_\theta$
    \ENDFOR
\ENDFOR
\RETURN $\mathcal{B}$
\end{algorithmic}
\end{algorithm}

\begin{algorithm}[H]
\caption{Cubic Sparsity Scheduling}
\label{alg:cubic_schedule}
\begin{algorithmic}[1]
\REQUIRE Current epoch $t$
\REQUIRE Warmup epochs $t_{warmup} = 2$
\REQUIRE Total epochs $t_{total} = 20$
\REQUIRE Minimum sparsity $s_{min} = 0.3$
\REQUIRE Maximum sparsity $s_{max} = 0.7$
\ENSURE Target sparsity $s(t)$

\IF{$t < t_{warmup}$}
    \RETURN $s(t) = 0.0$
\ENDIF

\STATE $progress \leftarrow \frac{t - t_{warmup}}{t_{total} - t_{warmup}}$
\STATE $progress \leftarrow \min(1.0, \max(0.0, progress))$
\STATE $s(t) \leftarrow s_{min} + (s_{max} - s_{min}) \cdot progress^3$
\RETURN $s(t)$
\end{algorithmic}
\end{algorithm}

\begin{algorithm}[H]
\caption{Global Magnitude-Based Weight Selection}
\label{alg:global_magnitude}
\begin{algorithmic}[1]
\REQUIRE Binary masks $\mathcal{B}$
\REQUIRE Neural network modules $\mathcal{M}$
\REQUIRE Target sparsity $s_{target}$
\ENSURE Updated masks $\mathcal{B}'$

\STATE Initialize empty lists: $W_{active} \leftarrow []$, $P_{info} \leftarrow []$
\FOR{each module $M_i \in \mathcal{M}$}
    \FOR{each weight parameter $\theta \in M_i$}
        \STATE $B_\theta \leftarrow \mathcal{B}[\theta]$ \COMMENT{Get current mask}
        \STATE $I_{active} \leftarrow \text{indices where } B_\theta = 1$
        \STATE $W_{active\_weights} \leftarrow |\theta[I_{active}]|$ \COMMENT{Absolute magnitudes of active weights}
        \IF{$|W_{active\_weights}| > 0$}
            \STATE $W_{active}.\text{append}(W_{active\_weights})$
            \STATE $P_{info}.\text{append}((\theta, I_{active}))$
        \ENDIF
    \ENDFOR
\ENDFOR

\IF{$W_{active}$ is empty}
    \RETURN $\mathcal{B}$
\ENDIF

\STATE $W_{all} \leftarrow \text{concatenate}(W_{active})$
\STATE $n_{total} \leftarrow \sum_{\theta, I} |\theta|$ for all $(\theta, I) \in P_{info}$
\STATE $n_{active} \leftarrow |W_{all}|$
\STATE $n_{target\_pruned} \leftarrow \lfloor s_{target} \cdot n_{total} \rfloor$
\STATE $n_{currently\_pruned} \leftarrow n_{total} - n_{active}$
\STATE $n_{additional\_prune} \leftarrow \max(0, n_{target\_pruned} - n_{currently\_pruned})$

\IF{$n_{additional\_prune} = 0$ OR $n_{additional\_prune} \geq n_{active}$}
    \RETURN $\mathcal{B}$
\ENDIF

\STATE $\tau \leftarrow \text{kthvalue}(W_{all}, n_{additional\_prune})$ \COMMENT{Find pruning threshold}

\FOR{each $(\theta, I_{active}) \in P_{info}$}
    \STATE $B'_\theta \leftarrow \mathcal{B}[\theta].\text{clone}()$
    \STATE $W_{prune\_mask} \leftarrow |\theta[I_{active}]| < \tau$
    \FOR{each position $pos$ in $I_{active}$ where $W_{prune\_mask}[pos] = \text{True}$}
        \STATE $B'_\theta[pos] \leftarrow 0$ \COMMENT{Prune weight}
    \ENDFOR
    \STATE $\mathcal{B}[\theta] \leftarrow B'_\theta$
\ENDFOR

\RETURN $\mathcal{B}$
\end{algorithmic}
\end{algorithm}

\begin{algorithm}[H]
\caption{Synaptic Pruning Training Integration}
\label{alg:training_integration}
\begin{algorithmic}[1]
\REQUIRE Training dataset $\mathcal{D}$
\REQUIRE Neural network model $f_\theta$
\REQUIRE Optimizer $\mathcal{O}$
\REQUIRE Loss function $\mathcal{L}$
\REQUIRE Pruning frequency $f_{prune} = 5$ batches
\REQUIRE Total epochs $E$
\ENSURE Trained model $f_{\theta^*}$ with sparse structure

\STATE Initialize synaptic pruning masks $\mathcal{B}$ \COMMENT{Algorithm \ref{alg:synaptic_init}}
\STATE $batch\_count \leftarrow 0$

\FOR{epoch $e = 1$ to $E$}
    \FOR{each batch $(X, y) \in \mathcal{D}$}
        \STATE $batch\_count \leftarrow batch\_count + 1$
        
        \STATE // Forward pass
        \STATE $\hat{y} \leftarrow f_\theta(X)$
        \STATE $loss \leftarrow \mathcal{L}(\hat{y}, y)$
        
        \STATE // Backward pass
        \STATE $\mathcal{O}.\text{zero\_grad}()$
        \STATE $loss.\text{backward}()$
        
        \STATE // Pruning update (every $f_{prune}$ batches after warmup)
        \IF{$e \geq t_{warmup}$ AND $batch\_count \bmod f_{prune} = 0$}
            \STATE $s_{target} \leftarrow \text{CubicSchedule}(e)$ \COMMENT{Algorithm \ref{alg:cubic_schedule}}
            \STATE $\mathcal{B} \leftarrow \text{GlobalMagnitudePruning}(\mathcal{B}, s_{target})$ \COMMENT{Algorithm \ref{alg:global_magnitude}}
        \ENDIF
        
        \STATE // Optimizer step
        \STATE $\mathcal{O}.\text{step}()$
        
        \STATE // Apply masks to ensure pruned weights remain zero
        \STATE $\text{ApplyMasks}(\theta, \mathcal{B})$
    \ENDFOR
\ENDFOR

\RETURN $f_{\theta^*}$
\end{algorithmic}
\end{algorithm}

\begin{algorithm}[H]
\caption{Apply Pruning Masks}
\label{alg:apply_masks}
\begin{algorithmic}[1]
\REQUIRE Model parameters $\theta$
\REQUIRE Binary masks $\mathcal{B}$
\ENSURE Updated parameters with pruned connections set to zero

\FOR{each weight parameter $w \in \theta$}
    \IF{$w \in \mathcal{B}$}
        \STATE $B_w \leftarrow \mathcal{B}[w]$
        \STATE $w \leftarrow w \odot B_w$ \COMMENT{Element-wise multiplication to zero out pruned weights}
    \ENDIF
\ENDFOR
\end{algorithmic}
\end{algorithm}

\begin{algorithm}[H]
\caption{Sparsity Statistics Computation}
\label{alg:sparsity_stats}
\begin{algorithmic}[1]
\REQUIRE Binary masks $\mathcal{B}$
\REQUIRE Neural network modules $\mathcal{M}$
\ENSURE Sparsity statistics $\mathcal{S}$

\STATE Initialize statistics dictionary $\mathcal{S} \leftarrow \{\}$
\STATE $layer\_idx \leftarrow 0$

\FOR{each module $M_i \in \mathcal{M}$}
    \FOR{each weight parameter $\theta \in M_i$}
        \STATE $B_\theta \leftarrow \mathcal{B}[\theta]$
        \STATE $n_{total} \leftarrow |B_\theta|$ \COMMENT{Total number of weights}
        \STATE $n_{pruned} \leftarrow \sum_{j}(B_\theta[j] = 0)$ \COMMENT{Number of pruned weights}
        \STATE $sparsity \leftarrow \frac{n_{pruned}}{n_{total}}$
        
        \STATE $key \leftarrow M_i.\text{class\_name} + "\_" + \theta.\text{name} + "\_" + layer\_idx$
        \STATE $\mathcal{S}[key] \leftarrow \{$
        \STATE \quad $'total\_weights': n_{total},$
        \STATE \quad $'pruned\_weights': n_{pruned},$
        \STATE \quad $'sparsity': sparsity$
        \STATE $\}$
    \ENDFOR
    \STATE $layer\_idx \leftarrow layer\_idx + 1$
\ENDFOR

\RETURN $\mathcal{S}$
\end{algorithmic}
\end{algorithm}

\subsection{Biological Inspiration}

\noindent This approach mirrors biological synaptic pruning processes observed in neural development, where initially overproduced synaptic connections are selectively eliminated based on their functional importance \cite{zhu2017b, chechik1999, chklovskii2004}. The gradual nature of our cubic scheduling function reflects the progressive refinement seen in biological neural networks, where pruning occurs over extended periods rather than through abrupt elimination. Additionally, the global magnitude-based selection mechanism ensures that only the most functionally relevant connections are preserved, similar to how biological systems maintain synapses that contribute most effectively to information processing and learning.

\subsection{Comparison with Alternative Approaches}

\noindent Unlike alternative dropout approaches \cite{kingma2015, gal2017}, which introduce learnable parameters and stochastic behavior, our method provides deterministic, magnitude-based pruning that results in genuinely sparse networks. This approach differs from methods like SNIP \cite{lee2019} by operating during training rather than at initialization, allowing for dynamic adaptation to learned representations. The global comparison strategy ensures more balanced pruning across layers compared to layer-wise approaches, leading to more efficient overall network architectures. 

\subsection{Implementation}

\begin{figure}[h!]
\centering
\includegraphics[height=0.4\textheight, angle=90]{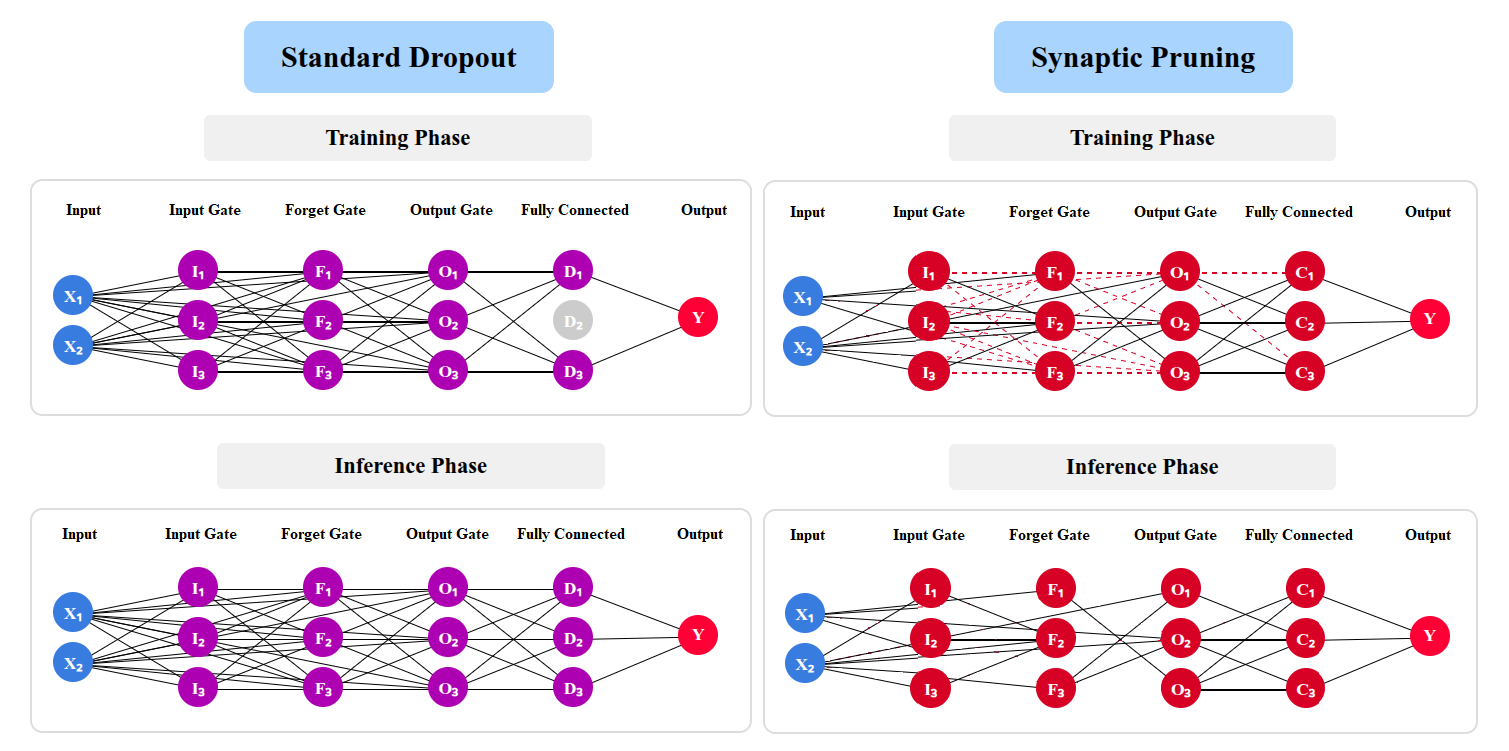}
\caption{\label{fig:figure1} Comparison of standard dropout and our synaptic pruning method implemented on LSTM model architectures. Dropout temporarily deactivates neurons during training, whereas pruning permanently removes specific connections, resulting in lasting sparsity and improved efficiency.}%
\end{figure}

\noindent When implemented on an LSTM architecture (Figure \ref{fig:figure1}), standard dropout randomly deactivates neurons in the hidden layers during training (indicated by gray nodes) while maintaining full network connectivity across all layers. During inference, all neurons are reactivated with no dropout applied. In contrast, our proposed synaptic pruning method permanently removes specific synaptic connections (indicated by red lines) during training, creating a progressively sparser network topology from across all layers.\\

\FloatBarrier

\begin{figure}[h!]
\centering
\includegraphics[height=0.4\textheight, angle=90]{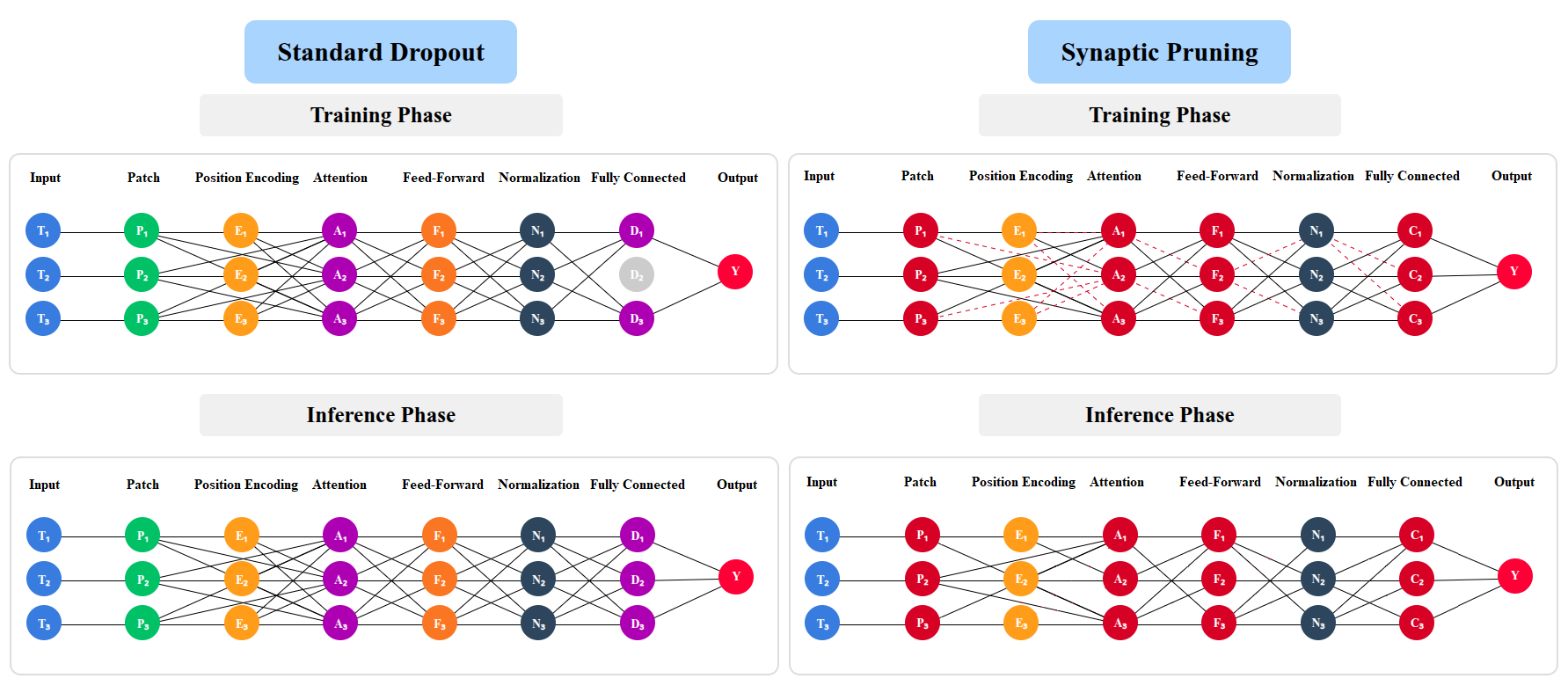}
\caption{\label{fig:figure2} Comparison of standard dropout and our synaptic pruning method implemented on PatchTST model architectures. Dropout temporarily deactivates neurons during training, whereas pruning permanently removes specific connections, resulting in lasting sparsity and improved efficiency.}%
\end{figure}

\noindent The transformer-based PatchTST implementation (Figure \ref{fig:figure2}) demonstrates similar principles but applies synaptic pruning from the patch to fully connected layers while standard dropout temporarily deactivates nodes within the fully connected layer (indicated by gray nodes). \\

\noindent Across both architectures, solid circles represent active neurons, light gray circles represent temporarily inactive neurons (dropout only), black lines indicate active connections, and red lines highlight pruned connections that remain permanently severed. The key distinction is that dropout provides temporary stochastic regularization without structural modification, while our proposed method creates permanent architectural sparsity that persists through inference, resulting in improved computational efficiency and reduced model complexity while maintaining performance.

\FloatBarrier

\subsection{Datasets}

\noindent This study employed four diverse public time-series datasets (Table \ref{tab:datasets}) to evaluate the proposed methodology across different domains and complexity levels. The datasets span various application areas including financial markets, cryptocurrency, energy consumption, and environmental monitoring, providing a comprehensive testbed for validation. \\

\noindent Two financial datasets were incorporated: a single stock ticker from the S\&P 500 Stocks dataset \cite{dataset_sp500} containing 1,259 records with 20 features, and the Bitcoin Daily historical dataset \cite{dataset_bitcoin} with 365 records and 7 features. Both financial datasets are classified as low complexity, representing well-structured time series data with established patterns typical of financial markets. \\

\noindent The Household Electric Power Consumption dataset \cite{dataset_power} provides the largest sample size with over 2 million records (2,075,259) and 9 features, offering extensive temporal data for energy usage analysis. Despite its substantial size, it maintains low complexity due to the straightforward nature of power consumption measurements. \\

\noindent Finally, the Air Quality dataset \cite{dataset_air} presents a moderate-sized dataset with 9,446 records but the highest feature dimensionality at 313 features, resulting in high complexity. This dataset challenges our proposed methodology with high-dimensional environmental data encompassing multiple pollutants and atmospheric conditions. The diversity in dataset characteristics, ranging from 365 to over 2 million records, 7 to 313 features, and varying complexity levels, ensures robust evaluation of the proposed method across different data scenarios and computational challenges.

\begin{table}[!h]
\centering
\caption{\label{tab:datasets}Datasets utilized in this study.}
\resizebox{\textwidth}{!}{
\begin{tabular}{lrrc}
\hline\hline
\textbf{Dataset}                     & \textbf{Record Count} & \multicolumn{1}{l}{\textbf{Features}} & \textbf{Complexity}  \\
\hline
\rowcolor[rgb]{0.753,0.753,0.753} S\&P 500 Stocks \cite{dataset_sp500}                       & 1259                  & 20                                    & Low                  \\
Bitcoin Daily \cite{dataset_bitcoin}                   & 365                   & 7                                     & Low                  \\
\rowcolor[rgb]{0.753,0.753,0.753} Household Electric Power Consumption \cite{dataset_power} & 2075259               & 9                                     & Low                  \\
Air Quality \cite{dataset_air}                          & 9446                  & 313                                   & High        \\                            
\hline
\end{tabular}
}
\end{table}
\FloatBarrier

\subsection{Experimentation}
\noindent A total of 96 experiments were performed using RNN, LSTM, and PatchTST architectures, measuring Mean Absolute Error (MAE) when using no regularization, standard dropout, Monte Carlo dropout \cite{gal2016} and our proposed pruning method, using input sequence lengths of 1, 3, 7, 14, 30 and 60. Each experiment was run for 10 trials using varying random seed initialization across a maximum of 20 epochs per trial, with the means of MAE and 95\% Confidence Interval (CI) calculated across trials. Experimentation was performed using an Intel i9 20-core CPU and a single Nvidia RTX2080ti GPU. The Python programming language version 3.13.2 with PyTorch version 2.6 was used for all experimentation, with the full source code and utilized data publicly hosted at https://github.com/xalentis/SynapticPruning.

\section{Results and Discussion}

\noindent Appendix A contains the full results of the 96 experiments. The provided source code includes all experimentation, feature-engineering functions and results analysis modules.

\subsection{Application to Recurrent Neural Networks (RNN)}

\begin{figure}[h!]
\centering
\includegraphics[width=\textwidth]{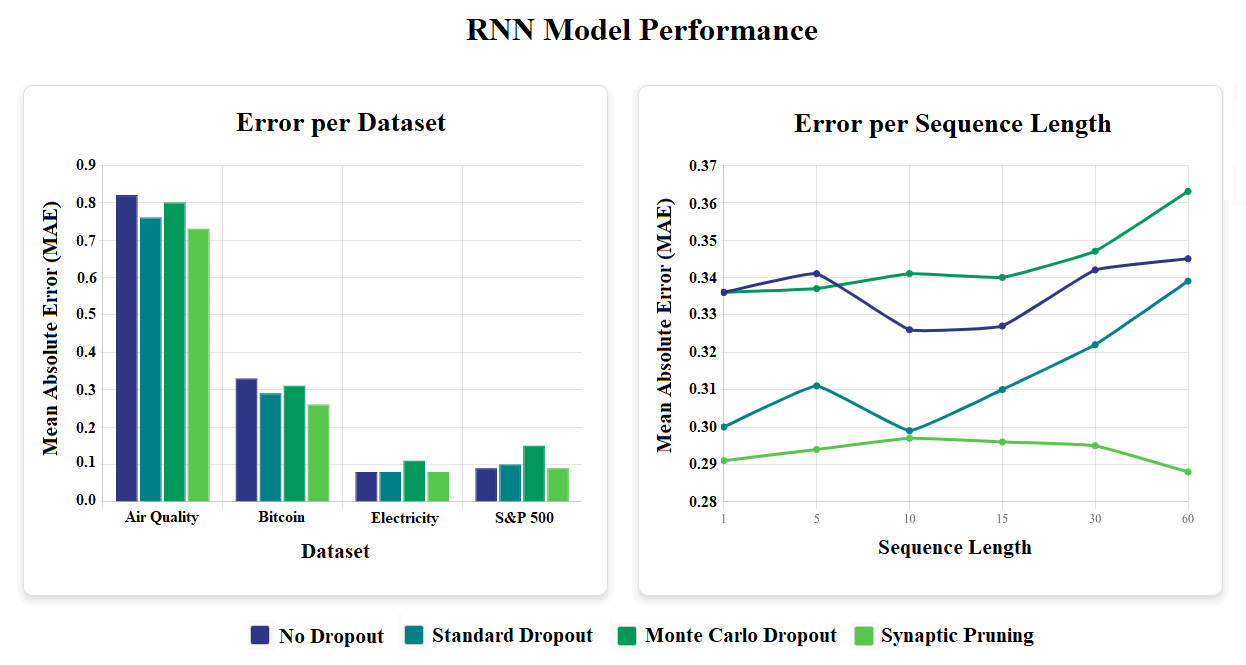}
\caption{\label{fig:figure3} Error rate comparison of regularization methods implemented on RNN model architectures.}%
\end{figure}

\noindent The RNN experimental results (Table \ref{tab:results_rnn} and Figure \ref{fig:figure3}) confirms that our proposed pruning method consistently outperforms Monte Carlo Dropout across all datasets, with statistically significant improvements in MAE (p \textless 0.05). Compared to standard Dropout, Synaptic Pruning delivers significant gains for the Bitcoin Daily, S\&P 500, and Air Quality datasets, but shows no significant difference on the Household Electric Power Consumption dataset. When compared to models trained without regularization, our method achieves comparable or better performance, with statistically significant improvements observed for the Bitcoin, Air Quality, and Electricity datasets, but not for the S\&P500 dataset. In terms of computational cost, our method incurs an average overhead of approximately 4.4\%, which is modest given its consistent performance gains.

\begin{table}[htbp]
\centering
\caption{\label{tab:results_rnn}RNN model experimentation results and statistical significance.}
\resizebox{\textwidth}{!}{
\begin{tabular}{|l|l|c|c|c|c|c|}
\hline
\textbf{Dataset} & \textbf{Method} & \textbf{Mean MAE} & \textbf{Std Dev} & \textbf{95\% CI} & \textbf{Friedman Test} & \textbf{Overall Significance} \\
\hline
\multirow{4}{*}{\textbf{Bitcoin Daily}} 
& No Dropout & 0.3296 & 0.0346 & [0.2933, 0.3659] & $\chi^2 = 15.80$ & \textbf{p = 0.001246*} \\
& Dropout & 0.2969 & 0.0443 & [0.2504, 0.3434] & & \\
& MC Dropout & 0.3119 & 0.0315 & [0.2788, 0.3450] & & \\
& \textbf{Synaptic Pruning} & \textbf{0.2621} & \textbf{0.0183} & \textbf{[0.2429, 0.2813]} & & \\
\hline
\multirow{4}{*}{\textbf{S\&P500}} 
& No Dropout & 0.0944 & 0.0156 & [0.0780, 0.1108] & $\chi^2 = 16.76$ & \textbf{p = 0.000792*} \\
& Dropout & 0.1105 & 0.0107 & [0.0993, 0.1217] & & \\
& \textbf{MC Dropout} & \textbf{0.1535} & \textbf{0.0082} & \textbf{[0.1449, 0.1621]} & & \\
& Synaptic Pruning & 0.0944 & 0.0156 & [0.0780, 0.1108] & & \\
\hline
\multirow{4}{*}{\textbf{Air Quality}} 
& No Dropout & 0.8298 & 0.0219 & [0.8068, 0.8528] & $\chi^2 = 16.40$ & \textbf{p = 0.000939*} \\
& Dropout & 0.7569 & 0.0356 & [0.7195, 0.7943] & & \\
& MC Dropout & 0.7992 & 0.0308 & [0.7669, 0.8315] & & \\
& \textbf{Synaptic Pruning} & \textbf{0.7265} & \textbf{0.0273} & \textbf{[0.6978, 0.7552]} & & \\
\hline
\multirow{4}{*}{\textbf{Household Electric Power Consumption}} 
& No Dropout & 0.0902 & 0.0043 & [0.0856, 0.0947] & $\chi^2 = 14.69$ & \textbf{p = 0.002097} \\
& Dropout & 0.0876 & 0.0046 & [0.0828, 0.0924] & & \\
& MC Dropout & 0.1108 & 0.0034 & [0.1073, 0.1144] & & \\
& \textbf{Synaptic Pruning} & \textbf{0.0855} & \textbf{0.0036} & \textbf{[0.0816, 0.0893]} & & \\
\hline
\end{tabular}
}
\end{table}
\FloatBarrier

\FloatBarrier

\subsection{Application to Long Short-Term Memory (LSTM)}

\begin{figure}[h!]
\centering
\includegraphics[width=\textwidth]{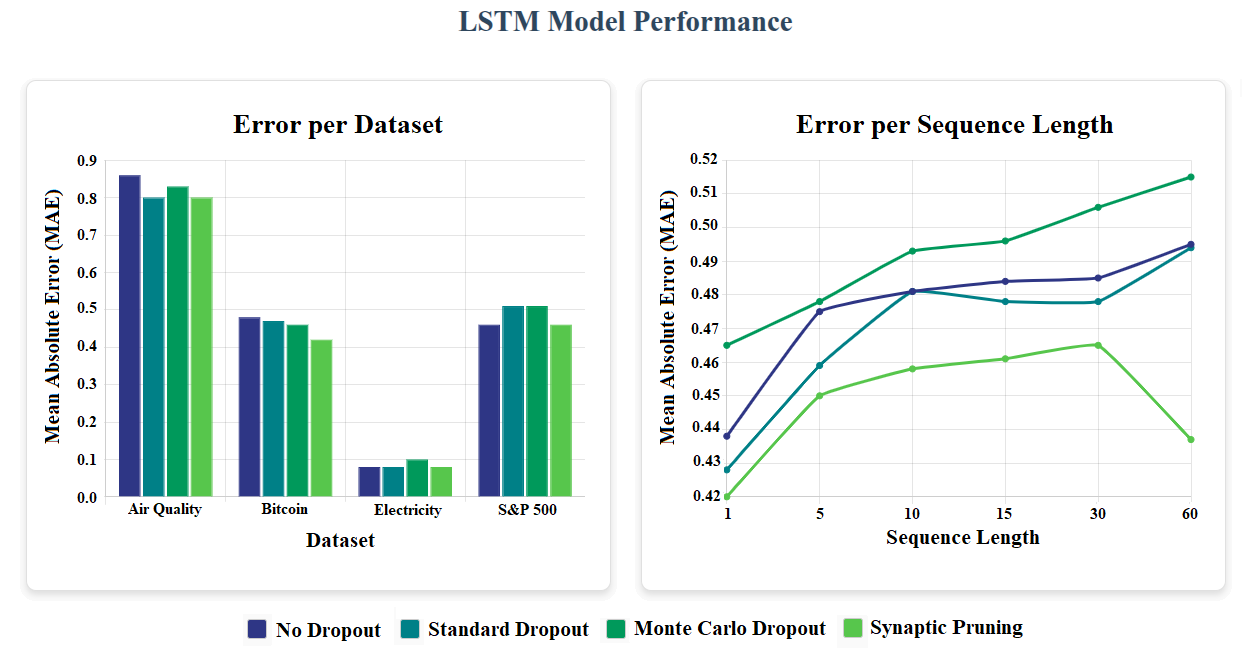}
\caption{\label{fig:figure4} Error rate comparison of regularization methods implemented on LSTM model architectures.}%
\end{figure}

\noindent LSTM experimental results (Table \ref{tab:results_lstm} and Figure \ref{fig:figure4}) reveals that our proposed method achieves an average MAE reduction of 9.8\% across all datasets, with performance degradation occurring in 3.3\% of cases. Performance gains vary notably by dataset, with the S\&P500 dataset exhibiting the largest improvement with a 21.5\% average MAE reduction, while the Household Electric Power Consumption dataset shows moderate gains of 8.3\%. In contrast, the Bitcoin Daily and Air Quality datasets display marginal improvements of 2.8\% and 2.4\%, respectively, with limited statistical significance. These findings suggest that while the method's effectiveness can vary depending on the dataset characteristics, it consistently offers meaningful improvements in optimal scenarios with minimal risk of adverse impact, supporting its general applicability for time series forecasting tasks using LSTM models.\\

\begin{table}[htbp]
\centering
\caption{\label{tab:results_lstm}Long Short-Term Memory experimentation results.}
\resizebox{\textwidth}{!}{
\begin{tabular}{|l|l|c|c|c|c|c|}
\hline
\textbf{Dataset} & \textbf{Method} & \textbf{Mean MAE} & \textbf{Std Dev} & \textbf{95\% CI} & \textbf{Friedman Test} & \textbf{Overall Significance} \\
\hline
\multirow{4}{*}{Bitcoin Daily} & No Dropout & 0.4804 & 0.1332 & [0.3406, 0.6202] & $\chi^2 = 12.80$ & \textbf{p = 0.005090} \\
 & Dropout & 0.4666 & 0.1281 & [0.3321, 0.6010] & & \\
 & MC Dropout & 0.4629 & 0.1208 & [0.3361, 0.5897] & & \\
 & \textbf{Synaptic Pruning} & \textbf{0.4212} & \textbf{0.1083} & \textbf{[0.3076, 0.5348]} & & \\
\hline
\multirow{4}{*}{S\&P500} & No Dropout & 0.4629 & 0.0914 & [0.3670, 0.5589] & $\chi^2 = 15.40$ & \textbf{p = 0.001505} \\
 & Dropout & 0.5060 & 0.0703 & [0.4322, 0.5798] & & \\
 & MC Dropout & 0.5102 & 0.0692 & [0.4376, 0.5828] & & \\
 & \textbf{Synaptic Pruning} & \textbf{0.4626} & \textbf{0.0915} & \textbf{[0.3667, 0.5586]} & & \\
\hline
\multirow{4}{*}{Air Quality} & No Dropout & 0.8634 & 0.0394 & [0.8220, 0.9048] & $\chi^2 = 15.20$ & \textbf{p = 0.001653} \\
 & Dropout & 0.8030 & 0.0415 & [0.7595, 0.8466] & & \\
 & MC Dropout & 0.8431 & 0.0408 & [0.8002, 0.8859] & & \\
 & \textbf{Synaptic Pruning} & \textbf{0.8107} & \textbf{0.0409} & \textbf{[0.7677, 0.8537]} & & \\
\hline
\multirow{4}{*}{Household Electric Power Consumption} & No Dropout & 0.0814 & 0.0022 & [0.0791, 0.0836] & $\chi^2 = 11.60$ & \textbf{p = 0.008887} \\
 & Dropout & 0.0813 & 0.0015 & [0.0797, 0.0829] & & \\
 & MC Dropout & 0.1046 & 0.0011 & [0.1035, 0.1058] & & \\
 & \textbf{Synaptic Pruning} & \textbf{0.0808} & \textbf{0.0022} & \textbf{[0.0784, 0.0831]} & & \\
\hline
\end{tabular}
}
\end{table}
\FloatBarrier

\FloatBarrier

\subsection{Application to Transformer Networks (PatchTST)}

\begin{figure}[h!]
\centering
\includegraphics[width=\textwidth]{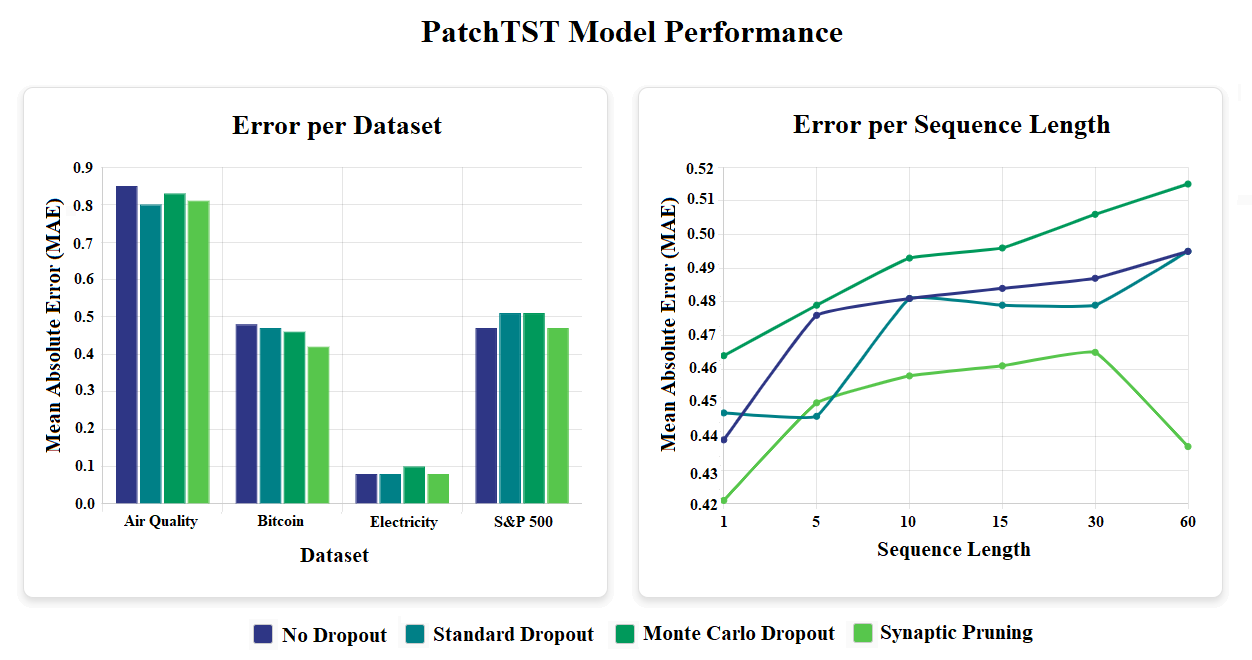}
\caption{\label{fig:figure5} Error rate comparison of regularization methods implemented on PatchTST model architectures.}%
\end{figure}

\noindent Experimental results using the PatchTST architecture (Figure \ref{fig:figure5}, Table \ref{tab:results_patchtst}) shows our proposed method achieving the best average MAE at 0.42, significantly outperforming using no regularization (0.52), dropout (0.55), and Monte Carlo methods (0.56), with our method reducing MAE by 17.5\% to 24.1\%. Our proposed method shows statistically significant improvements in 62.5\% of experiments, with no cases where it performs significantly worse than alternative methods, and 37.5\% showing no significant difference. Performance of our method varies by dataset, with Bitcoin Daily showing exceptional results at 32.0\% improvement over other methods, while on the SP\&500 dataset demonstrates more modest gains of 10.3\% improvement. \\

\noindent Longer sequence lengths benefit more from our pruning method, with 60-step sequences showing 22.0\% improvement to 30-step sequences which demonstrate 16.7\% improvement. Our proposed method did not perform worse than competing methods across any configuration, potentially establishing it as a reliable regularization technique that consistently maintains or improves model performance.

\begin{table}[htbp]
\centering
\caption{\label{tab:results_patchtst}PatchTST experimentation results.}
\resizebox{\textwidth}{!}{
\begin{tabular}{|l|l|c|c|l|c|l|}
\hline
\textbf{Dataset} & \textbf{Method} & \textbf{Mean MAE} & \textbf{Std MAE} & \textbf{Friedman Test} & \textbf{Rank} & \textbf{Best Pairwise Comparisons (p-value)} \\
\hline
\multirow{4}{*}{\textbf{Bitcoin}} 
& No Dropout & 0.3710 & 0.1481 & \textbf{p=0.000939*} & 2 & vs Dropout (p=0.0312*), vs MC Dropout (p=0.0312*) \\
& Dropout & 0.7740 & 0.1413 &  & 4 & Significantly worse than all others \\
& MC Dropout & 0.4620 & 0.1175 &  & 3 & vs Dropout (p=0.0312*), vs Synaptic (p=0.0312*) \\
& \textbf{Synaptic Pruning} & \textbf{0.3692} & 0.1460 &  & \textbf{1} & \textbf{Best performer} \\
\hline
\multirow{4}{*}{\textbf{S\&P500}} 
& No Dropout & 0.3952 & 0.1086 & \textbf{p=0.003071*} & 3 & vs MC Dropout (p=0.0312*) \\
& \textbf{Dropout} & \textbf{0.3312} & 0.0685 &  & \textbf{1} & vs MC Dropout (p=0.0312*) \\
& MC Dropout & 0.5923 & 0.0604 &  & 4 & Significantly worse than No Dropout and Dropout \\
& Synaptic Pruning & 0.3123 & 0.0538 &  & 2 & vs MC Dropout (p=0.0312*) \\
\hline
\multirow{4}{*}{\textbf{Air Quality}} 
& No Dropout & 17.6980 & 2.4818 & \textbf{p=0.003847*} & 3 & vs Dropout (p=0.0312*) \\
& Dropout & 23.4991 & 2.7051 &  & 4 & Significantly worse than MC Dropout and Synaptic \\
& \textbf{MC Dropout} & \textbf{16.4987} & 0.9511 &  & \textbf{1} & vs Dropout (p=0.0312*) \\
& Synaptic Pruning & 16.8684 & 2.2254 &  & 2 & vs Dropout (p=0.0312*) \\
\hline
\multirow{4}{*}{\textbf{Household Electric Power Consumption}} 
& \textbf{No Dropout} & \textbf{0.1102} & 0.0093 & \textbf{p=0.000440*} & 2 & vs Dropout (p=0.0312*), vs MC Dropout (p=0.0312*) \\
& Dropout & 0.3114 & 0.0208 &  & 4 & Significantly worse than all others \\
& MC Dropout & 0.1316 & 0.0067 &  & 3 & vs Dropout (p=0.0312*), vs Synaptic (p=0.0312*) \\
& \textbf{Synaptic Pruning} & \textbf{0.1060} & 0.0083 &  & \textbf{1} & \textbf{Best performer} \\
\hline
\end{tabular}
}
\end{table}
\FloatBarrier

\FloatBarrier

\subsection{Computational Efficiency and Network Sparsification}

\begin{figure}[h!]
\centering
\includegraphics[width=\textwidth]{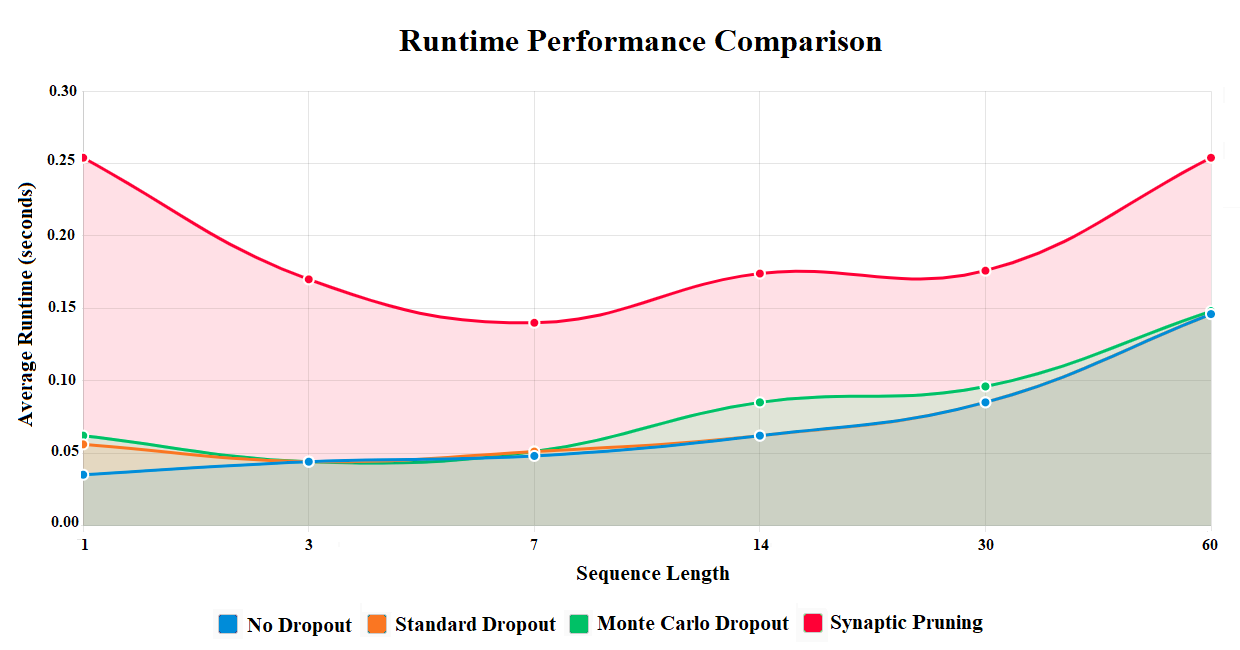}
\caption{\label{fig:figure6} Average peak memory usage vs sequence length by model".}%
\end{figure}

\noindent The proposed synaptic pruning method demonstrates a positive trade-off between computational efficiency and performance across varying sequence lengths. The runtime performance comparison (Figure \ref{fig:figure6}) reveals that while synaptic pruning initially exhibits higher computational overhead at short sequences, it achieves efficiency gains at longer sequences. At sequence lengths 1 and 60, synaptic pruning reaches its optimal performance (Figure \ref{fig:figure5}), significantly outperforming both standard dropout and Monte Carlo dropout (which shows gradual degradation). However, the method exhibits an inverted U-shaped performance curve, though still maintaining competitive performance compared to the baseline without dropout. Where computational costs do increase, these are justified by substantial improvements in predictive accuracy, with the method consistently achieving lower error rates than baseline approaches, demonstrating that the additional computational investment delivers meaningful performance returns.\\

\FloatBarrier

\subsection{Biological Plausibility and Neuro-scientific Validation}

\noindent Our proposed method's biological inspiration extends beyond superficial analogies to incorporate quantifiable aspects of synaptic pruning mechanisms, mirroring activity-dependent synaptic strengthening observed in long-term potentiation (LTP) \cite{bliss1993} and long-term depression (LTD) \cite{bear1996}. The temporal dynamics of our pruning process align with developmental timescales observed in cortical maturation \cite{bourgeois1997}. This biological grounding distinguishes our approach from purely engineering-based pruning methods and suggests potential for further bio-inspired enhancements. \\

\noindent Performance results reveal that synaptic pruning is particularly well-suited for moderate sequence lengths (3-14), where it achieves optimal efficiency. This behavior pattern aligns with the adaptive nature of biological synaptic pruning, where neural circuits optimize themselves for frequently encountered input patterns. The method's ability to maintain stable memory usage while providing variable runtime performance could be particularly valuable for applications with diverse sequence length requirements \cite{howard2017, iandola2016}. Our analysis shows that networks converge to optimal sparsity levels within the first 15 epochs, after which the pruning rate stabilizes. This early convergence behavior aligns with biological observations of critical periods of brain development \cite{hensch2004, takesian2013}, suggesting that our method captures fundamental principles of neural circuit optimization during these developmental windows.

\section{Study Limitations}

\noindent While the results from the study indicate potential in the use of a more targeted approach to regularization compared to current dropout approaches, a number of limitations require future work and experimentation. First, a relatively small number of model architectures and network configurations were tested, limiting generalizability to other critical architectures such as vision transformers, and modern large language models. Secondly, experimentation was conducted using CPU or a single GPU system without using parallel data transfer during the pruning phase, which will require additional consideration when experimenting on significantly larger training datasets. The computational overhead observed may compound significantly at enterprise scale, and our limited experimentation may not reflect modern big data scenarios where scalability becomes critical.

\section{Conclusion and Future Work}

\noindent This study introduces a biologically-inspired regularization technique that aims to mimic synaptic pruning in the developing brain, offering an alternative to current dropout approaches for neural network regularization. By dynamically evaluating neurons based on importance metrics and permanently removing connections with minimal contribution to error reduction, our proposed approach significantly enhances predictive accuracy. The error reduction observed in our PatchTST experiments, when applied to time series forecasting, demonstrates the efficacy of activity-dependent pruning over conventional regularization methods.\\

\noindent The success of our approach aligns with the neuro-scientific understanding of brain development, where synaptic elimination occurs based on functional utility rather than stochastic processes. This bio-mimetic strategy preserves critical pathways while removing redundant connections, resulting in sparser networks with improved generalization capabilities. \\

\noindent Future work should address the computational overhead through optimized implementation and parallelization techniques with validation on larger datasets. Long-term research directions should explore implementations that include both synaptic growth and pruning phases, potentially leading to next-generation bio-inspired architectures that fundamentally change how neural networks learn and adapt. By continuing to incorporate neuro-biological principles into deep learning, we envision improved and novel techniques that enable neural networks to better emulate the remarkable efficiency and generalization capabilities of biological learning systems. To support further research in this area, the full source code used in this study is available on GitHub at https://github.com/xalentis/SynapticPruning

\section{Supplementary Material: Appendix A}
\includepdf[pages=-]{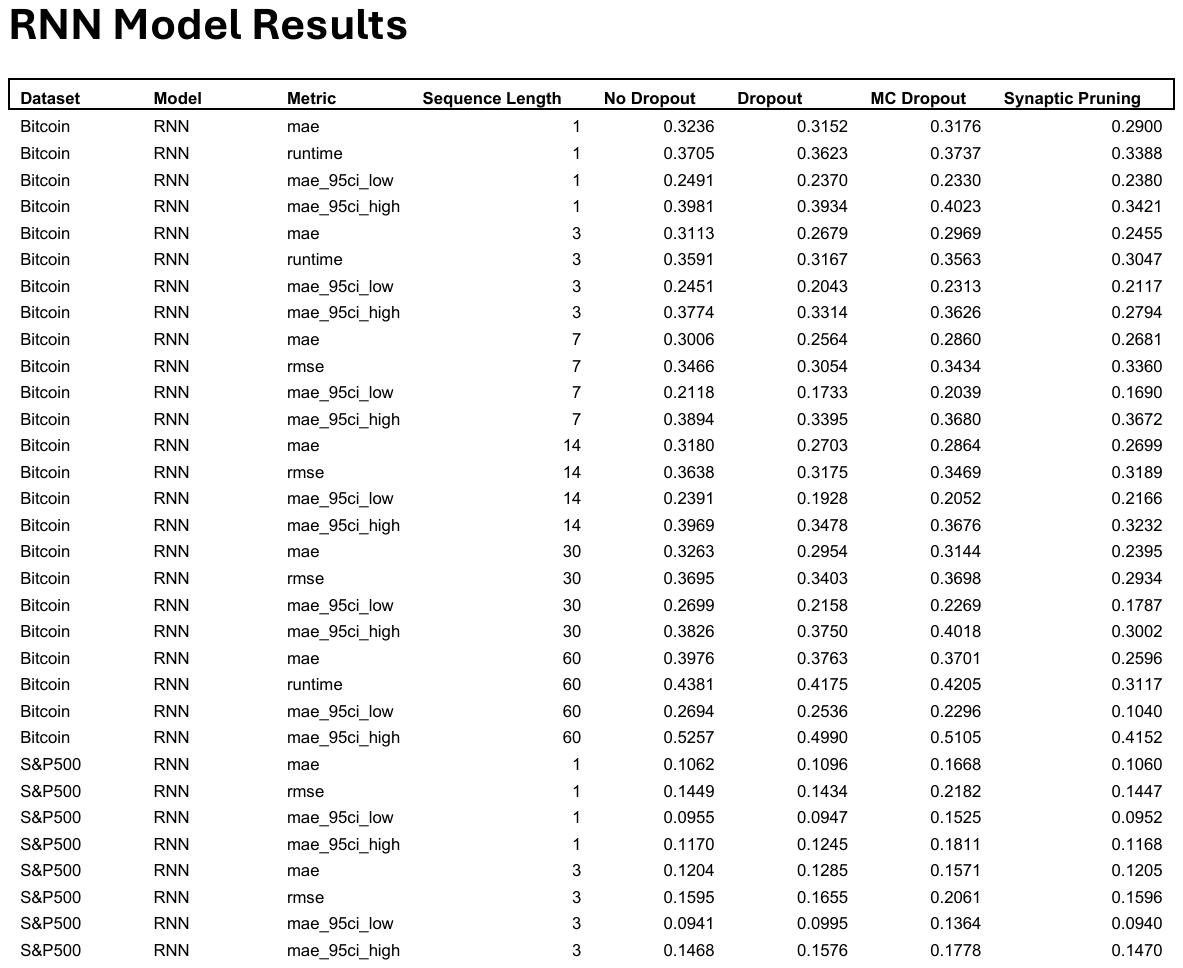}

 \bibliographystyle{elsarticle-num} 
 \bibliography{cas-refs}






\end{document}